\documentclass[letterpaper, 10 pt, journal, twoside]{ieeetran} 

\IEEEoverridecommandlockouts                              

\usepackage{cite}
\usepackage{amsfonts}
\usepackage{algorithmic}
\usepackage{graphicx}
\usepackage{textcomp}

\usepackage{amsmath}
\usepackage{amssymb}

\usepackage{amsthm}
\usepackage{graphicx}
\usepackage{mathtools}

\newtheorem*{assumption}{Assumption}
\usepackage[hidelinks]{hyperref}
\usepackage{colortbl}
\usepackage{caption}
\usepackage{subcaption}
\usepackage{balance}
\usepackage[linesnumbered,ruled,vlined]{algorithm2e}

\SetCommentSty{mycommfont}
\SetKwInput{KwInput}{Input}
\SetKwInput{KwOutput}{Output}
\usepackage{siunitx}
\usepackage{hyperref}

\usepackage[most]{tcolorbox}
\usepackage{bm}

\title{\LARGE \bf
Infinite-Horizon Value Function Approximation\\ for Model Predictive Control
}

\author{Armand Jordana$^{1}$, Sébastien Kleff$^{1,4}$, Arthur Haffemayer$^{3}$, Joaquim Ortiz-Haro$^{1}$, \\Justin Carpentier$^{2}$, Nicolas Mansard$^{3, 5}$ and Ludovic Righetti$^{1, 5}$
\thanks{Manuscript received: January 31, 2025; Revised: April 28, 2025; Accepted: May 29, 2025.}
\thanks{
This paper was recommended for publication by Editor Lucia Pallottino upon evaluation of the Associate Editor and Reviewers’comments.}
\thanks{This work was in part supported by the National Science Foundation grants 1932187, 2026479, 2222815 and 2315396 and ANITI (ANR 19-P3IA-0004). This work was supported by a grant overseen by the French National Research Agency (ANR) and France 2030 as part of the PR[AI]RIE-PSAI AI cluster (ANR-23-IACL-0008), by the French government under the management of Agence Nationale de la Recherche through the project NIMBLE (ANR-22-CE33-0008), and by the  European Union through the AGIMUS project (GA no.101070165).}
\thanks{\textsuperscript{1}~Machines in Motion Laboratory, New York University, USA}%
\thanks{\textsuperscript{2}~Inria - Département d’Informatique de l’École normale supérieure, PSL Research University.}
\thanks{\textsuperscript{3}\, LAAS-CNRS, Universit{\'e} de Toulouse, CNRS, Toulouse, France}%
\thanks{\textsuperscript{4}\, Inria, AUCTUS team, Talence, France}%
\thanks{\textsuperscript{5}\, Artificial and Natural Intelligence Toulouse Institute (ANITI), Toulouse}%
\thanks{\textsuperscript{$\star$}\, Corresponding author: \tt\small armand.jordana@nyu.edu}
\thanks{Digital Object Identifier 10.1109/LRA.2025.3577875}
}%

\begin{document}

\IEEEpubid{\begin{minipage}{\textwidth}\ \\[10pt]
\centering
© 2025 IEEE. Personal use of this material is permitted. Permission from IEEE must be obtained for all other uses, in any current or future media, including reprinting/republishing this material for advertising or promotional purposes, creating new collective works, for resale or redistribution to servers or lists, or reuse of any copyrighted component of this work in other works.
\end{minipage}}

\maketitle

\begin{abstract}
Model Predictive Control has emerged as a popular tool for robots to generate complex motions. However, the real-time requirement has limited the use of hard constraints and large preview horizons, which are necessary to ensure safety and stability. In practice, practitioners have to carefully design cost functions that can imitate an infinite horizon formulation, which is tedious and often results in local minima. In this work, we study how to approximate the infinite horizon value function of constrained optimal control problems with neural networks using value iteration and trajectory optimization. Furthermore, we experimentally demonstrate how using this value function approximation as a terminal cost provides global stability to the model predictive controller. The approach is validated on two toy problems and a real-world scenario with online obstacle avoidance on an industrial manipulator where the value function is conditioned to the goal and obstacle.
\end{abstract}

\begin{IEEEkeywords}
Optimization and Optimal Control, Machine Learning for Robot Control.
\end{IEEEkeywords}
\section{INTRODUCTION}

Model Predictive Control (MPC) has demonstrated its ability to plan efficiently online on complex robots~\mbox{\cite{koenemannwhole2015, neunert2018whole, sleiman2021unified, dantec2022whole}}. However,  in robotics applications, MPC exhibits local behaviors for two main reasons. 
The first is that practitioners often rely on local Trajectory Optimization~(TO) techniques.
The second is the use of a finite horizon which creates local minima. 
Furthermore, without an appropriate design of the terminal cost and constraint set, MPC with a finite horizon is not guaranteed to be globally stable or recursively feasible~\cite{mayne2014model, grune2017nonlinear}.
In practice, this has led roboticists to spend a lot of time designing cost functions to avoid local minima. Infinite horizon constrained MPC is a compelling framework as it ensures global stability~\cite{grune2017nonlinear}. In other words, it guarantees that any initial state converges to a zero-cost stationary state while satisfying hard constraints. Unfortunately, in the general case, the problem is intractable and has to be approximated~\cite{bertsekas2019reinforcement}.

Reinforcement Learning (RL)~\cite{sutton2018reinforcement} and Approximate Dynamic Programming (ADP)~\cite{bertsekas2019reinforcement} appear as good candidates to move the compute time offline by approximating policies or value functions. Those techniques recently have shown impressive results in locomotion~\cite{agarwal2023legged, hoeller2024anymal} and manipulation~\cite{chen2023visual, handa2023dextreme}. However, they are subject to the curse of dimensionality and can be unsafe outside of the training distribution. Also, incorporating safety by imposing hard constraints is challenging with current RL and ADP tools. In contrast, by re-planning online, constrained MPC can adapt to novel situations and ensure hard constraint satisfaction.

Combining the advantages of online and offline decision making is appealing, and it has proven to be highly effective in the context of games such as Go or Chess~\cite{silver2016mastering}. In the context of robotics, it could allow maintaining safety by ensuring hard constraints while leveraging the full potential of function approximation. 
In this work, we propose to combine MPC with function approximation in order to perform infinite horizon constrained MPC. The infinite horizon value function is approximated using neural networks and trained using value iteration and local gradient-based optimization. Then, during deployment, the approximated value is used as a terminal cost for the Optimal Control Problem (OCP). We show experimentally that the approximated value function helps the controller escape local minima while online optimization compensates for approximation errors. Furthermore, outside of the training distribution, the controller might not be optimal but, online optimization ensures hard constraint satisfaction given a control-invariant set.

\IEEEpubidadjcol

The theoretical benefits of an infinite horizon formulation have been extensively studied~\cite{chen1998quasi, mayne2000constrained, hu2002toward, grune2017nonlinear}. Unfortunately, the general case is intractable. Hence, efforts have concentrated on the constrained linear quadratic regulator~\cite{bemporad2002explicit, grieder2004computation, stathopoulos2016solving}.
In the general case, RL provides a way to approximate the solution~\cite{sutton2018reinforcement, bertsekas2019reinforcement}. 
In the discrete action setting, Deep Q-learning is a popular tool~\cite{mnih2015human}. In the continuous case, actor-critics such as DDPG~\cite{lillicrap2015continuous} or SAC~\cite{haarnoja2018soft} allow to learn simultaneously a policy and a value function. However, despite recent progress~\cite{zhang2022penalized, chane2024cat}, incorporating hard constraints in those formulations remains challenging. Also, RL algorithms~\cite{sutton2018reinforcement} typically use a discount factor. However, the global stability guarantees of infinite horizon MPC were established in the non-discounted setting~\cite{mayne2014model, grune2017nonlinear}. Consequently, we study the non-discounted setting with hard constraints which, to the best of our knowledge, remains understudied in the RL community.

The idea of combining MPC and function approximation is not novel and has fostered a lot of research in the robotics community. The seminal work of Atkeson \cite{atkeson1993using} explored how to use local trajectory optimization together with a global value function. Since then, an extensive amount of work has shown the benefits of using TO with learning to either speed up the training of value functions and policies or improve the controller's performance at test time~\cite{zhong2013value, levine2013guided, mordatch2014combining, korda2016controller, lowrey2018plan, hoeller2020deep, landry2021seagul, hatch2021value, Parag2022, grandesso2023cacto, alboni2024cacto}. 
A limitation of these works is their inability to consider hard nonlinear constraints. In practice, constraints have to be enforced softly using penalty terms in the cost function. However, this approach requires tedious weight tuning and can hardly provide guarantees.
In this work, we show how the known advantages of combining MPC and RL can be obtained while enforcing hard constraints. 

More recently, \cite{viereck2022valuenetqp, wang2024online} demonstrated on hardware the benefits of using online constrained optimization with an approximate value as a terminal model in the context of locomotion. In these works, the authors propose to learn the value with a local TO solver using a long horizon. Consequently, at test time, the controller remains local. In contrast, we use value iteration combined with local TO to approximate the infinite horizon value function and we demonstrate the ability of the method to avoid local minima.

In this work, we propose to approximate the infinite horizon value function of constrained OCP and use it as a terminal cost function of a Model Predictive Controller. The contributions of this paper are threefold:
\begin{itemize}
    \item First, we demonstrate how a local gradient-based solver allows the use of value iteration to approximate the optimal value function of an infinite horizon constrained OCP. 
    \item Second, we provide an experimental study showing how the use of trajectory optimization can compensate for the inaccuracies of the value function approximation.
    \item Third, we demonstrate the benefits of combining MPC with value function approximation on a reaching task with obstacle avoidance on an industrial manipulator. More precisely, we show experimentally how the use of the value function allows avoiding the local minima MPC is subject to. Furthermore, given a control-invariant set, we demonstrate how the method remains safe by ensuring hard constraints outside the training distribution of the value function. To the best of our knowledge, this is the first demonstration of MPC using a learned infinite horizon value function with hard constraints deployed on a robot at real-time rates.
\end{itemize}

\section{BACKGROUND}
In this work, we are interested in the infinite-horizon constrained optimal control problem:
\begin{subequations}
\begin{align}
    V(x) = \lim_{T\to\infty}&  \min_{u_0, u_1, \dots u_{T-1}}\sum_{k=0}^{T-1} \ell (x_k, u_k) \\
    \mbox{  s.t.  } x_0 &= x\\
    x_{k+1} &= f(x_k, u_k) \\
    c(x_k, u_k) &\geq 0
\end{align} \label{main_definition}
\end{subequations}
Here $V$ is the infinite horizon value function. The state $x$ belongs to $\mathbb{R}^{n_x}$ and the control $u$ to $\mathbb{R}^{n_u}$. The dynamics function $f$ maps $\mathbb{R}^{n_x} \times \mathbb{R}^{n_u}$ to $\mathbb{R}^{n_x}$. The constraint function $c$ maps  $\mathbb{R}^{n_x} \times \mathbb{R}^{n_u}$ to $\mathbb{R}^{n_c}$.
The cost function $\ell$ maps $\mathbb{R}^{n_x} \times \mathbb{R}^{n_u}$ to  $\mathbb {R} ^{+}$. Similarly to \cite{bertsekas2015value}, to ensure the existence of states yielding a finite value function, we consider that:
\begin{assumption}
The set of stationary points yielding a zero cost 
\begin{align}
\mathcal{G} = \{ x \, | \, \exists \, u \, \text{s.t.} \, \ell(x, u) = 0, \, x = f(x, u), \, c(x, u) \geq 0 \}
\end{align}
is not empty.
\end{assumption}
 Due to the infinite sum, computing the value function or its associated optimal policy is intractable in general. However, RL provides tools to find an approximation.

Let's denote $\Omega$, the space on which the value function is well-defined. That is to say the set of initial states such that there exists a control sequence that ensures constraint satisfaction for all future states. For any state $x \in \Omega$, the value function satisfies the Bellman equation:
\begin{align}
       \mathcal{B} \left(V\right) = V,
\end{align}
where $\mathcal{B}$ is the Bellman operator, defined by:
\begin{subequations}
\begin{align}
     \mathcal{B}\left(V\right)(x) =  &\min_{u}  \ell (x, u) + V(f(x, u))  \\
    \mbox{  s.t.  } c(x, u) &\geq 0\\
    f(x, u) &\in \Omega \label{eq:rec_feasibility}
\end{align}
\label{eq:bellman}
\end{subequations}
Note that by definition, $\Omega$ is a control-invariant set. Hence, Eq~\eqref{eq:rec_feasibility} ensures recursive feasibility~\cite{grune2017nonlinear} and safety~\cite{ames2019control}. In this work, we focus on problems where the set $\Omega$ is known and can be expressed analytically, which encompasses many problems. For instance, if the constraint is of the form $c(u)$, then $\Omega = \mathbb{R}^{n_x}$. For a fully actuated system, if the constraint is of the form $c(x)$ where $x$ denotes position and velocity and $u$ denotes torque control inputs, then ${\Omega = \{ x \in \mathbb{R}^{n_x} | c(x) \geq 0\}}$. In the more general case where the dynamics are nonlinear and the constraint is a function of both the state and the control, the set~$\Omega$ is not tractable, and we would have to rely on approximation techniques~\cite{djeridane2006neural, bansal2021deepreach}. However, this is beyond the scope of the paper. 

In the non-discounted setting, the Bellman equation generally has multiple solutions since adding a positive constant to any solution produces
another solution~\cite{bertsekas2015value}. However, the fact that points in $G$ yield a zero cost can allow us to recover the value function defined in Eq~\eqref{main_definition}. In fact, a necessary and sufficient condition to recover the value function defined in Eq~\eqref{main_definition} is to find a function that satisfies the Bellman equation and that yields a zero value on stationary points~\cite{bertsekas2015value}. 
In other words, we need to find a function $V$ satisfying the Bellman equation~\eqref{eq:bellman} and such that, 
\begin{align}
  \forall x \in  \mathcal{G}, \quad  V(x) = 0.  \label{eq:V_stat}
\end{align}
Value iteration provides a way to find a solution to the Bellman equation by iteratively applying the Bellman operator:
\begin{align}
    V_{k+1} = \mathcal{B}(V_k). \label{eq:value_iteration}
\end{align}
Under the assumption that the set $\mathcal G$ is not empty, it can be shown that $V_k$ converges to $V$ pointwise~\cite{heydari2014revisiting, bertsekas2015value}.

\section{METHOD}

\subsection{$T$-step optimal lookahead problem}

In this work, we propose to combine online and offline decision-making by using the $T$-step optimal lookahead problem~\cite{bertsekas2012dynamic} online. The idea is to perform MPC by solving an OCP at each time step, using an approximate value function as a terminal cost function.
More precisely, given a state $x$, we aim to find the optimal action by solving:
\begin{subequations} \label{main_eq}
\begin{align}
     \min_{u_0, u_1, \dots u_{T-1}}& \sum_{k=0}^{T-1} \ell (x_k, u_k) + V_{\theta}(x_{T})  \\
    \mbox{  s.t.  } x_0 &= x\\
    x_{k+1} &= f(x_k, u_k) \\
    c(x_k, u_k) &\geq 0\\
    x_T &\in \Omega
\end{align}
\end{subequations}
Here, $V_{\theta}$ is the value function approximation. In this work, we consider $V_{\theta}$ to be a neural network parameterized by weights $\theta$.
At each time step, the first optimal control input, $u_0^{\star}$, is applied to the system, and the other controls are disregarded. In the end, the policy, $\pi^{\star}(x)=u_0^{\star}$, depends on the horizon~$T$ and the approximated value function $ V_{\theta}$. In continuous action space, RL algorithms rely on a function approximation of the policy~\cite{sutton2018reinforcement}. In contrast, we solve Problem~\eqref{main_eq} online. The benefit of this approach is that the model used during the first $T$ steps of the optimization can reduce the inaccuracies of the value function and guarantee stability~\cite{bertsekas2005dynamic, krener2021adaptive, bertsekas2024model}. Furthermore, given knowledge of~$\Omega$, this ensures hard constraint satisfaction despite the use of an approximated value function.

\subsection{Value function approximation}

In this section, we show how to use value iteration defined in Equation~\eqref{eq:value_iteration} to approximate the value function of an infinite-horizon constrained OCP by neural networks. To minimize the Bellman Equation~\eqref{eq:bellman}, we propose to use a local gradient-based solver. To reduce the required number of value iterations (i.e. iterations of the Bellman operator), we apply the minimization over an arbitrary horizon $T$. More precisely, we use trajectory optimization to directly solve:
\begin{align}
    V_{k+1}  = \mathcal{B}^{[T]} \left(V_k\right) \label{eq:multiple_bellman}
\end{align}
where $\mathcal{B}^{[T]} $ is the Bellman operator over a horizon~$T$. 
\begin{align}
 \mathcal{B}^{[T]} \left(V\right)  = \underbrace{ \mathcal{B} \circ \hdots \circ \mathcal{B}}_{T~\text{times}}\left(V\right)
\end{align}
Indeed, it can be shown by induction that iterating $T$ times the Bellman operator is equivalent to solving an OCP of horizon $T$, precisely as in Problem~\eqref{main_eq}. Intuitively, this should allow us to perform $T$ times fewer value iterations.

In this work, we propose to fit the approximated value function at each Bellman iteration in a supervised way. Note that this approach can be considered as an instance of Fitted Value Iteration (FVI)~\cite{munos2008finite, bertsekas2019reinforcement} adapted to the constrained and deterministic setting.
More precisely, at iteration~$k$, given a value function $V_k$, we sample $n$ states, $\{x_{j}\}_{1\leq j \leq n}$ and solve the $n$ associated OCP with the corresponding initial condition and $V_k$ as a terminal cost. Then, we train in a supervised way such that $V_{\theta}$ maps $x_j$ to $\mathcal{B}^{[T]} \left(V_k\right)(x_j)$. Furthermore, to ensure that $\forall x^s \in \mathcal{G}, V_{\theta}(x^s) = 0$, we sample $m$ stationary points, $\{x^s_{j}\}_{1\leq j \leq m}$, and minimize the value function at those points with the Mean Squared Error (MSE). In the end, we minimize the following loss:
\begin{align}
\sum_{j=1}^n \left( V_{\theta}(x_j) -  \mathcal{B}^{[T]} \left(V_k\right)(x_j) \right)^2 + \alpha \sum_{j=1}^m   V_{\theta}(x^s_{j})^2,  \label{eq:main_loss}
\end{align}
where $\alpha$ is a penalty parameter. 
To create the targets, we solve Problem~\eqref{main_eq} using the stagewise Sequential Quadratic Programming (SQP) implementation introduced in~\cite{jordana2023stagewise}. This solver can handle hard constraints and exploits the time sparsity of the problem by using Riccati recursions in order to guarantee a linear complexity with the time horizon and quadratic convergence. Algorithm~\ref{algo_fitted_value_iteratoin} summarizes the method.

\begin{algorithm}[!ht]
\DontPrintSemicolon
\KwInput{dynamics $f$, cost $\ell$, constraint $c$, horizon $T$, network parameters $\theta$}
 Initialize $V_1$\;
\tcc{Main value iteration loop} 
\For{$k \gets 1$ to $N$}{
\tcc{Generate data} 
\For{$j \gets 1$ to $M$}{
Sample $x_j$\;
Compute $\mathcal{B}^{[T]} \left(V_k\right)(x_j)$ by optimizing \eqref{main_eq}\;
}
Create dataset:  $ \mathcal{D} = \{ (x_j, \mathcal{B}^{[T]} \left(V_k\right)(x_j))_j\}$\;
\tcc{Fit value function with SGD} 
\For{$j \gets 1$ to $P$}{
Sample batch from $\mathcal{D}$\;
Sample $m$ stationary points\;
Apply gradient descent with loss \eqref{eq:main_loss};
}
$V_{k+1} \gets V_{\theta}$
}
\KwOutput{$V_{N+1}$}
\caption{Value Iteration}\label{algo_fitted_value_iteratoin}
\end{algorithm}

Minimizing \eqref{main_eq} with a gradient-based solver requires the derivatives of the neural network. More specifically, the SQP approach requires the gradient and Hessian of the terminal cost function. To circumvent deriving twice a neural network, we use the Gauss-Newton approximation and define the value function as the squared L2 norm of a residual, i.e:
\begin{align}
    V_{\theta}(x) = \frac{1}{2} \| r_{\theta}(x) \|_2^2
\end{align}
where $r_{\theta}$ is a neural network with outputs in $\mathbb{R}^d$, where $d$ is a hyperparameter. Using the Gauss-Newton approximation, we have:
\begin{subequations}
\begin{align}
   \partial_x V_{\theta}(x) &= \partial_x r_{\theta}(x)^T r_{\theta}(x) \\
   \partial^2_{xx} V_{\theta}(x) &\approx \partial_x r_{\theta}(x)^T  \partial_x r_{\theta}(x)
\end{align}
\end{subequations}
where $\partial r_{\theta}(x) \in  \mathbb{R}^{d \times n_x}$ is the Jacobian matrix of  $r_{\theta}$ evaluated in $x$. This formulation is computationally efficient and also has the advantage of encoding the positiveness of the value function. 
For numerical optimization to be performed efficiently, it is crucial to obtain an accurate Jacobian of the value function. While \cite{Parag2022} investigated the use of Sobolev learning \cite{czarnecki2017sobolev}, we found that using \textit{tanh} activation functions and an appropriate weight decay was sufficient to ensure convergence of the SQP in few iterations.
Furthermore, we find that initializing the initial guess on the value function, $V_1$, to the zero function improves the training. To do so, at the first value iteration, we solve the OCP without a terminal cost function.

In order to generate trajectories that are similar to the one encountered at deployment, we rollout trajectories to generate more data. More specifically, after solving Problem~\eqref{main_eq}, the first state of the trajectory, $x_1$, is added to the dataset by solving Problem~\eqref{main_eq} using $x_1$ as an initial condition, and we iterate until either the goal or a maximum number of iterations is reached. In other words, Problem~\eqref{main_eq} is used as an MPC controller to collect additional data.
We find this especially relevant in complex examples with high-dimensional state space.

\section{EXPERIMENTS}

In this section, we present three problems of increasing complexity. The first two examples are used to illustrate the ability of the method to approximate the infinite horizon problem with a finite horizon and an approximate terminal value function. Lastly, we study a reaching task on an industrial manipulator with an obstacle to demonstrate the scalability of the method. In order to handle a moving scene, the value function is conditioned on the goal of the reaching task and the obstacle pose.
First, we provide an analysis of the impact of the horizon both at train and test time. Then, we present results from real experiments.
For all experiments, we use the SQP implementation introduced in \cite{jordana2023stagewise}. During the training, we solve each OCP in parallel on the CPU. Note that this is crucial to obtain reasonable training times. 

\subsection{Toy Problem 1: Constrained Simple Pendulum}

The first test problem we consider is the swing-up of a simple pendulum with torque limits. We illustrate how a finite horizon with an approximate value function as a terminal cost allows us to approximate the infinite horizon MPC.
The state is $x = \begin{bmatrix} \theta, \dot \theta \end{bmatrix}^T$ where $\theta$ denotes the orientation of the pendulum. The dynamics are defined by applying Euler integration to the following law of motion:
\begin{align}
    \ddot \theta = - \frac{g}{L} \sin(\theta)
\end{align}
The goal is to bring the pendulum to the upward position, which is incentivized with the following cost:
\begin{align}
    \ell(x, u) = \cos(\theta) + 1 + 0.01 \dot \theta ^ 2 + 0.001 u ^2   
\end{align}
Lastly, the control input, $u$, is constrained to be within $[-2, 2]$ which makes it impossible to swing up the pendulum without several back and forth.
We sample $\theta$ uniformly in $[-\pi, \pi]$ and $\dot \theta$ uniformly in $[-6, 6]$.
The number of sample points at each value iteration, $n$, is set to $500$, and the number of goals sampled $m$ is set to $1$ as $\mathcal{G}$ is a singleton. We consider a horizon of length $T=10$ and perform $1000$ value iterations. At each iteration, we perform $80$ SGD steps with $\alpha = 1$ using Adam with default parameters and a weight decay of $10^{-4}$. The network is a three-layer MLP with $64$ neurons and output size $d=64$. In the end, the training lasts 2 minutes and 40 seconds.

\begin{figure}[h!]
    \centering
    \includegraphics[width=\linewidth]{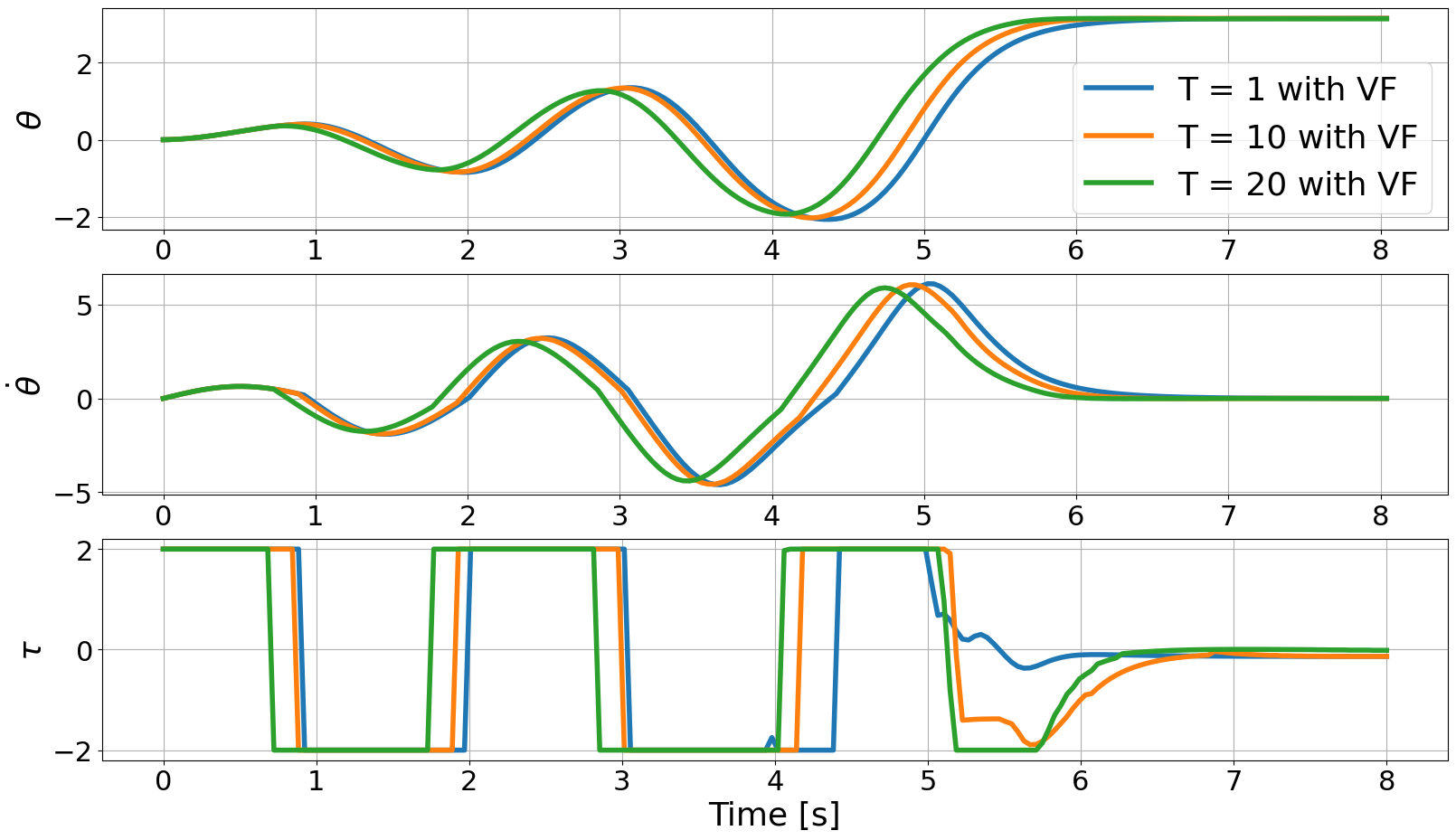}
    \caption{Rollout of MPC controllers with different horizon lengths using the learned value function as a terminal cost.}
    \label{fig:pendulum-trajectory}
\end{figure}

Figure~\ref{fig:pendulum-trajectory} shows the behavior of the MPC controller for different horizon lengths using the approximated value function as a terminal cost. The quality of the control increases as the horizon length increases. This can also be seen by looking at the final cumulative cost. The longer the horizon, the more optimal the controller is. For $T=1$, the cost is $4.57$, for $T=10$, the cost is $ 4.50$ and for $ T = 20$, the cost is $4.41$. This illustrates how solving online Problem~\eqref{main_eq} compensates for the approximation error of the value function.
\subsection{Toy Problem 2: Constrained point}
\begin{figure}[h!]
     \centering
     \begin{subfigure}[b]{0.49\linewidth}
         \centering
         \includegraphics[width=\linewidth]{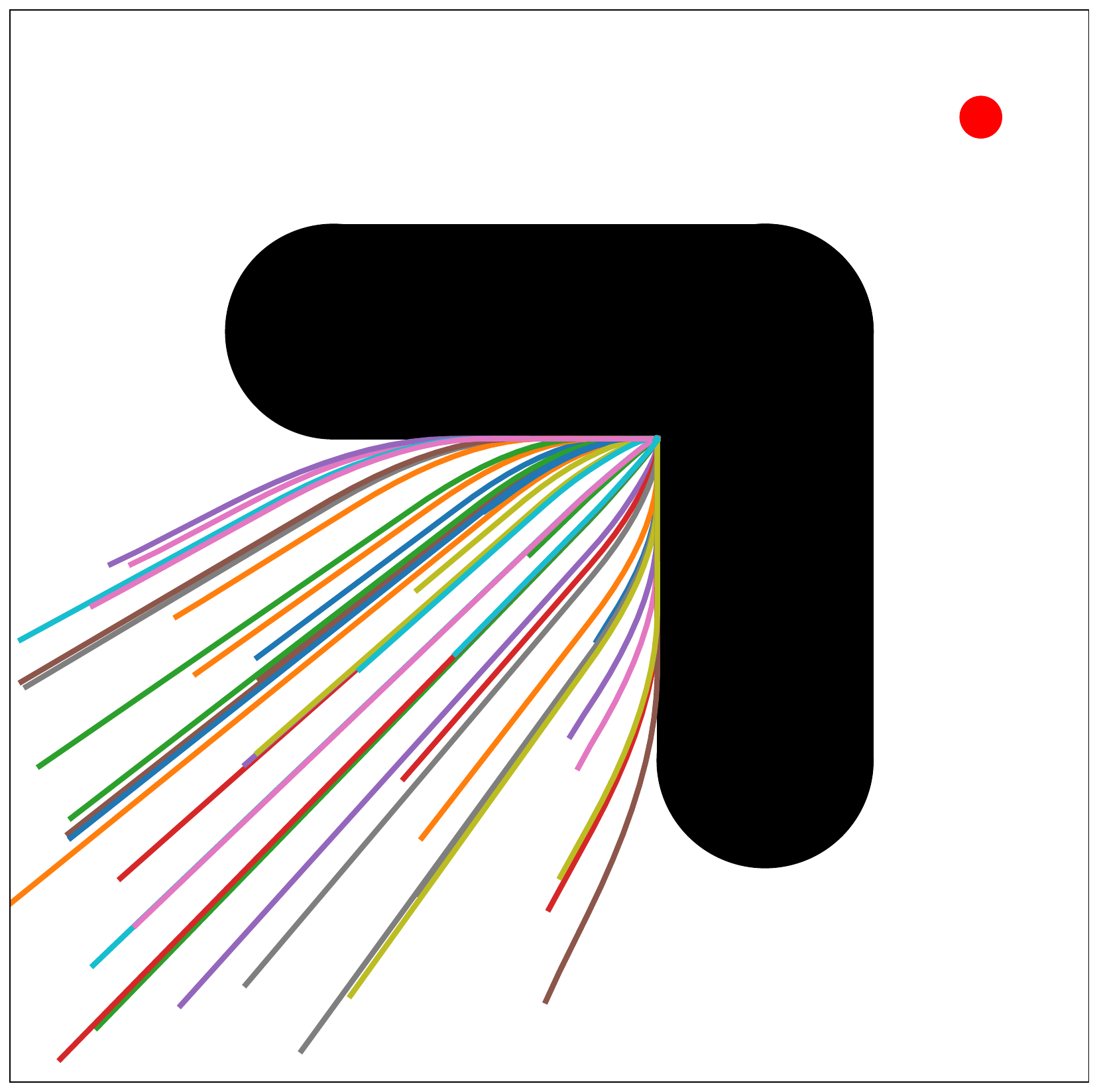}
         \caption{Without value function.}
     \end{subfigure}
     \begin{subfigure}[b]{0.49\linewidth}
         \centering
         \includegraphics[width=\linewidth]{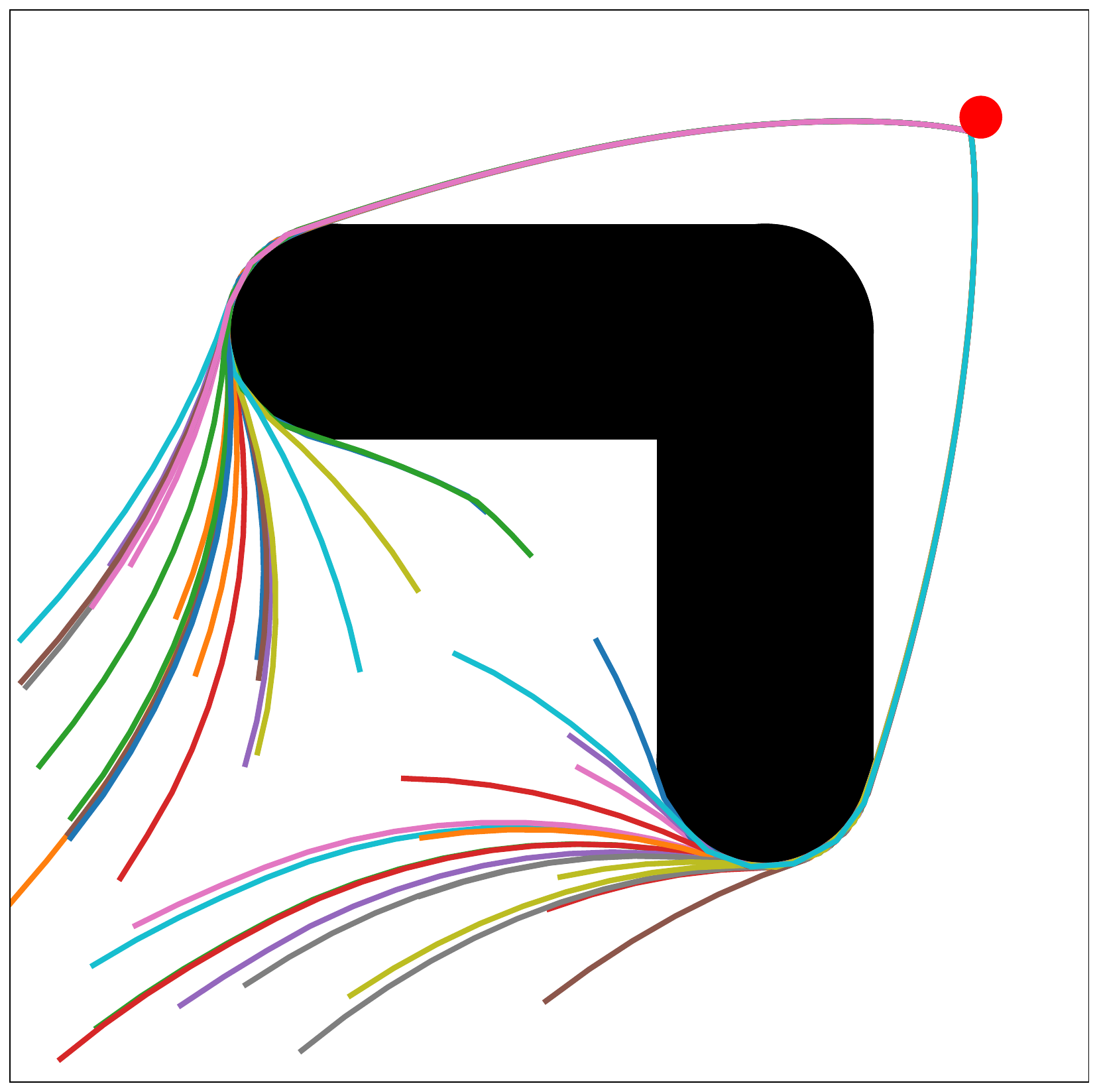}
         \caption{With value function.}
     \end{subfigure}
    \caption{MPC trajectories for different initial conditions. The red dot represents the target $x^{\star}$.}
    \label{fig:point}
\vspace{-0.4cm}
\end{figure}

In this second toy example, we illustrate the ability of the method to avoid local minima to which MPC is prone. We consider a 2-dimensional point that has to move around an obstacle to reach a target. The state, $x$, denotes the 2D position, and the control $u$ denotes the velocity of $x$. The dynamics and cost are defined in the following way:
\begin{align}
    x_{t+1}    &= x_t + \Delta t u_t\\
    \ell(x, u) &= \| x - x^{\star} \|_2^2 + 0.1 \| u \|_2^2
\end{align}
where $\Delta t$ is set to $0.02$. The constraints are defined by the distance between the point and the obstacle, as illustrated in Figure~\ref{fig:point}. We sample $x$ uniformly and reject states inside the obstacle.
The number of samples, $n$, is set to $2500$, and we augment the dataset with the last state of each trajectory. We chose $m = 1$ as $\mathcal{G}$ is a singleton. We consider a horizon of $10$ and perform $100$ value iterations using $\alpha = 1$. At each iteration, we perform $2000$ SGD steps using Adam with default parameters and a weight decay of $10^{-4}$. 
The network is a three-layer MLP with $32$ neurons and output size $d=32$.
The training lasts 12 minutes. The time increase compared to the previous experiment is due to the implementation of the model in Python.
Figure~\ref{fig:point} shows how using the learned value function as a terminal cost allows to bypass the obstacle to reach the goal. 
Without the value function, the controller gets stuck in the corner of the obstacle.

\subsection{Manipulator experiment: influence of the horizon}

\paragraph{At train time}

\begin{figure}[h!]
    \centering
    \includegraphics[width=\linewidth]{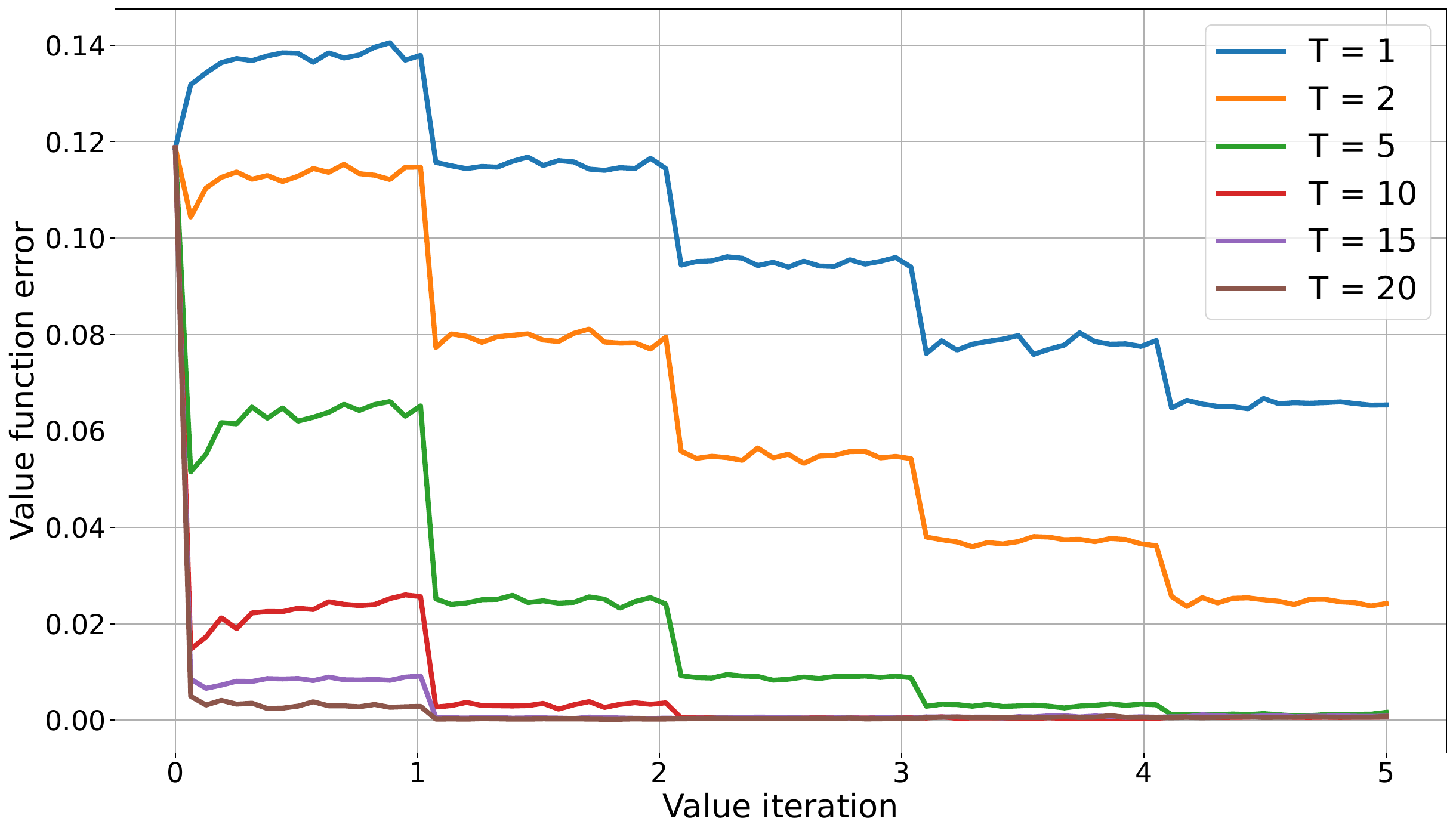}
    \caption{Error between the ground truth and the learned value function during training for various horizon length. The larger the horizon is, the faster the algorithm converges.}
    \label{fig:horizon_train}
\vspace{-0.2cm}
\end{figure}

In this section, we study the influence of the horizon on the training of the value function. We consider a reaching task with the 7-DoF Kuka iiwa robot. We consider an unconstrained OCP as in that setting, we can approximate the infinite sum in Eq.~\eqref{main_definition} with a large horizon~$T$. Consequently, we use our solver with $T=200$ (and no terminal cost) to approximate the ground truth infinite horizon value function. We fix the last joint and consider a 12-dimensional state containing joint positions and velocities. The OCP includes an end-effector target reaching cost, joint velocity regularization, and joint torque regularization costs. At each value iteration, we collect $10000$ trajectories of length $10$ by sampling the initial configuration uniformly within the joint and velocity bounds of the robot. Then, we perform $16$ epochs. Figure~\ref{fig:horizon_train} shows the MSE between the learned value and the ground truth during training. The larger the horizon in Eq~\eqref{main_eq} is, the faster the algorithm converges to the ground truth. Note that the data generation time is negligible compared to the time required by SGD to fit the network. Therefore, the total training time is similar for all horizon lengths. 

\paragraph{At test time}

\begin{figure}[h!]
    \centering
    \includegraphics[width=\linewidth]{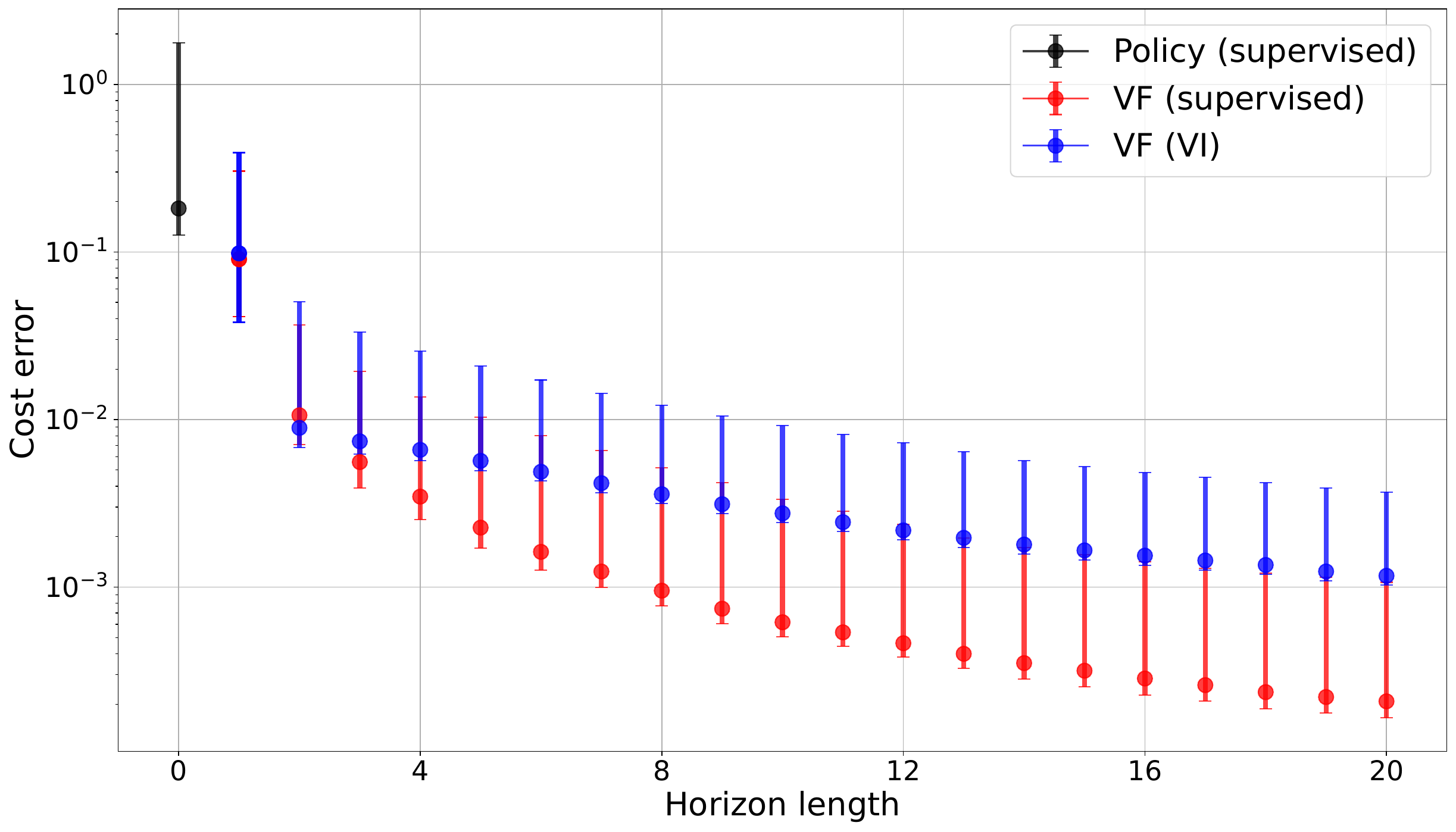}
    \caption{We run $1000$ MPC simulations starting from random initial states with increasing horizon for each controller. Horizon $0$ corresponds to the policy.}
    \label{fig:horizon_study_cost}
\vspace{-0.3cm}
\end{figure}

In this study, we show that online optimization can compensate for the value function approximation error due to learning. We use the same setup as in the previous section and show that at test time, a longer horizon helps to reduce the running cost. We perform value iteration with a horizon of~$10$ and use the same parameters as in the previous section. Furthermore, we illustrate that the improvement due to the horizon is not specific to our training procedure but due to the limitation of the neural network's expressivity. To do so, we train in a supervised way the value function with $100000$ ground truth MPC trajectories of length~$10$. We use the same number of SGD steps as in the overall value iteration learning procedure; therefore, the test time performance of this network can be considered as an upper bound on the one of the value iteration network.
Lastly, using the ground truth data, we train in a supervised way a policy mapping states to torques (while removing the gravity compensation). Figure~\ref{fig:horizon_study_cost} shows the cost error between various controllers and the ground truth infinite horizon controller. It can be seen that for both value functions, increasing the horizon improves performance. Also, value iteration training achieves a performance close to supervised training which is provided with the ground truth values.

\subsection{Real manipulator experiments}
\subsubsection{Setup}
We validate the proposed approach on the KUKA iiwa LBR 1480 in target reaching/tracking and obstacle avoidance tasks. We use a motion capture system (VICON) to track the targets and obstacles. The robot receives joint torque commands and a PD joint state reference at $\qty{1}{\kHz}$ through the FRI. The overall control law reads
\begin{align}
  \tau = u_0^{\star} - K_P (\hat{q} - q_1^{\star}) - K_D (\hat{\dot q} - {\dot q_1}^{\star})
\end{align}
where $u_0^{\star}$ and $x_1^{\star}$ are the optimal control input and the predicted state respectively (computed by the MPC at $\qty{100}{\Hz}$) and $\hat{x}$ is the measured state of joint positions and velocities, controlled at $\qty{1}{\kHz}$ by the joint PD. 
We use $K_P=[150, 150, 100, 100, 50, 10, 10]$ and $K_D = [25, 25, 20, 20, 14, 6, 6]$. The measured position is directly read from the robot's encoders, while the velocity is estimated by finite differences. Note that the last joint is blocked and not part of the model, as it speeds up learning and is not necessary for the tasks under study.

The obstacle we consider is a thin rod of length $\qty{76}{\cm}$. To enforce collision avoidance, we cover the robot with capsules and define the signed distance between the rod and the capsules, as well as between the table and the capsules, as hard constraints~\cite{haffemayer2024model}. In total, this represents $14$ constraints.

To deploy the method on the real system, we condition the network on both the target position and obstacle pose. States are sampled within $70\%$ of the robot's joint and velocity range. The target's Cartesian position is sampled within $[0.45, 0.75] \times [-0.2, 0.2] \times [0.15, 0.5]$ and states with end-effector positions outside this region are discarded, as these are outside the robot's workspace. Obstacle poses are sampled within a $\qty{10}{\cm}$ cube in front of the robot. We also randomize slightly the orientation through a uniform sampling of the Euler angles within $[-0.1, 0.1]$. 
Lastly, we reject the triplets (state, target, obstacle) with collisions between the robot and the obstacle or for which inverse kinematics has no solution for the given target. 
Additionally, to better reflect scenarios encountered on the real system, we reject triplets where both the target and end-effector are above the rod.
We consider $T=5$ and perform $500$ value iterations with $\alpha = 0.01$. At each iteration, we collect a dataset by sampling $1000$ triplets and rolling out trajectories up to $60$ time steps or until the robot reaches the target. The target is considered to be reached whenever the running cost is below $0.1$. Then, we perform  $16$ epochs using Adam with a learning rate of $0.0004$ and a weight decay of $10^{-5}$. The training lasts $\qty{3}{\hour}~\qty{20}{\minute}$ on CPU only.

During deployment, the learned value function is used as a terminal cost in the MPC, while the baseline MPC has no terminal cost. The horizon used is $T=10$, with an OCP discretization of $\Delta T = \qty{50}{\ms}$. The maximum number of SQP iterations is set to $6$, the termination tolerance to $10^{-4}$, and the maximum number of QP iterations to $200$. 
The OCP includes the non-collision constraints and state limits. 

We use a 4-layer MLP with $64$ neurons per layer and an output size $d=64$.
To minimize compute time, the network and its derivative are implemented in C++. For our architecture, the inference time is similar to the one of the model ($\qty{19}{\us}$ for the network and $\qty{16}{\us}$ for the model). In other words, the neural network does not limit the horizon length we can consider during real-time experiments. 

\subsubsection{Pick-and-place with static obstacle}
\begin{figure*}[t!]
\centering
\vspace{5px}
\begin{tabular}{c @{\hspace{1px}} c @{\hspace{1px}} c @{\hspace{1px}} c @{\hspace{1px}} c @{\hspace{1px}} c @{\hspace{1px}} c @{\hspace{1px}} c}
  \includegraphics[width=.12\textwidth, height=.12\textheight]{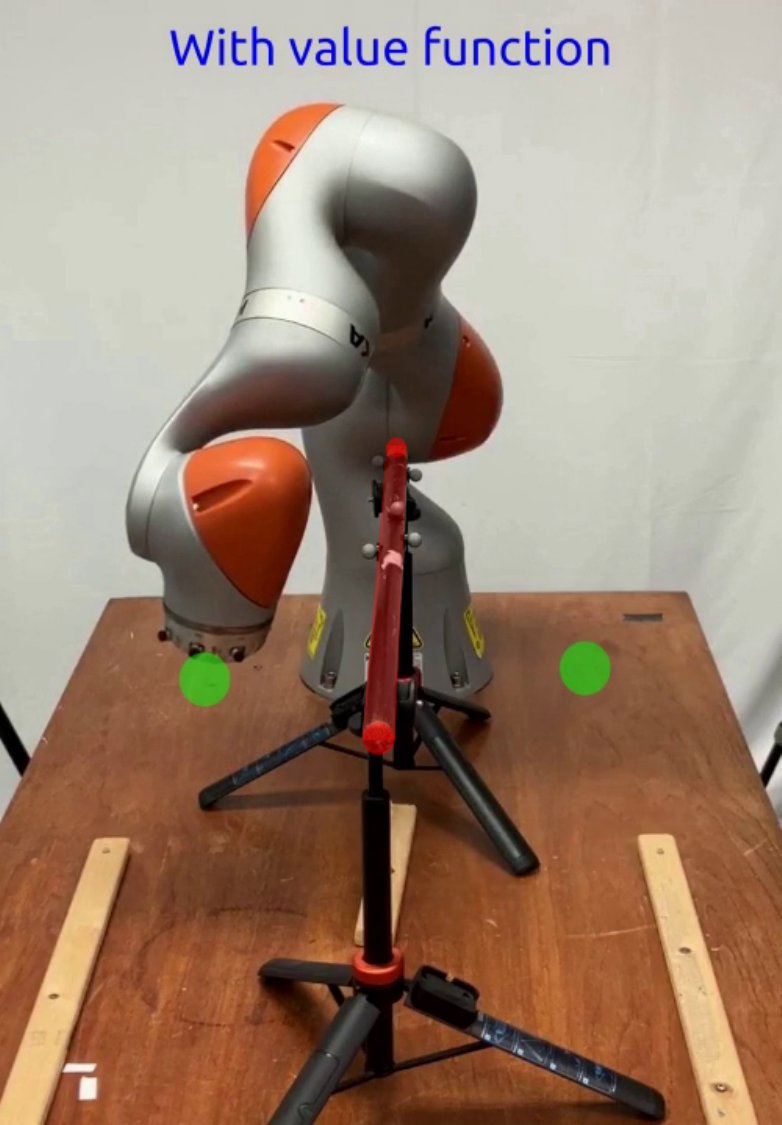}&
  \includegraphics[width=.12\textwidth, height=.12\textheight]{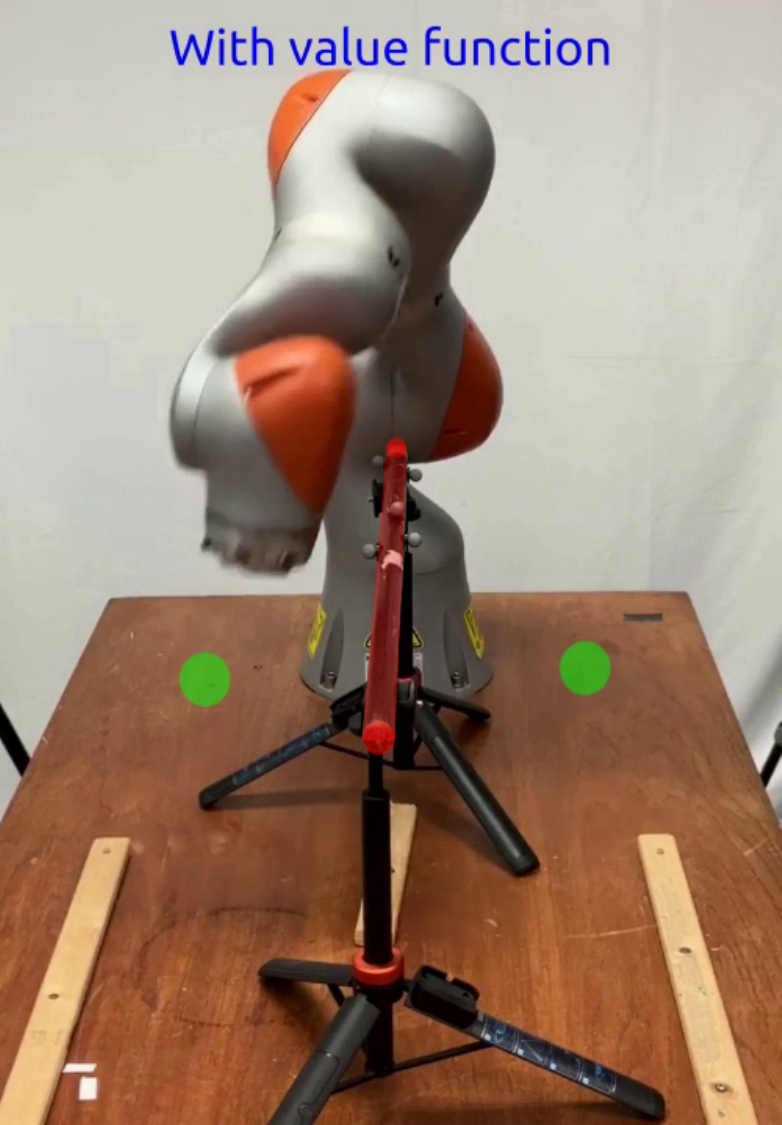}&
  \includegraphics[width=.12\textwidth, height=.12\textheight]{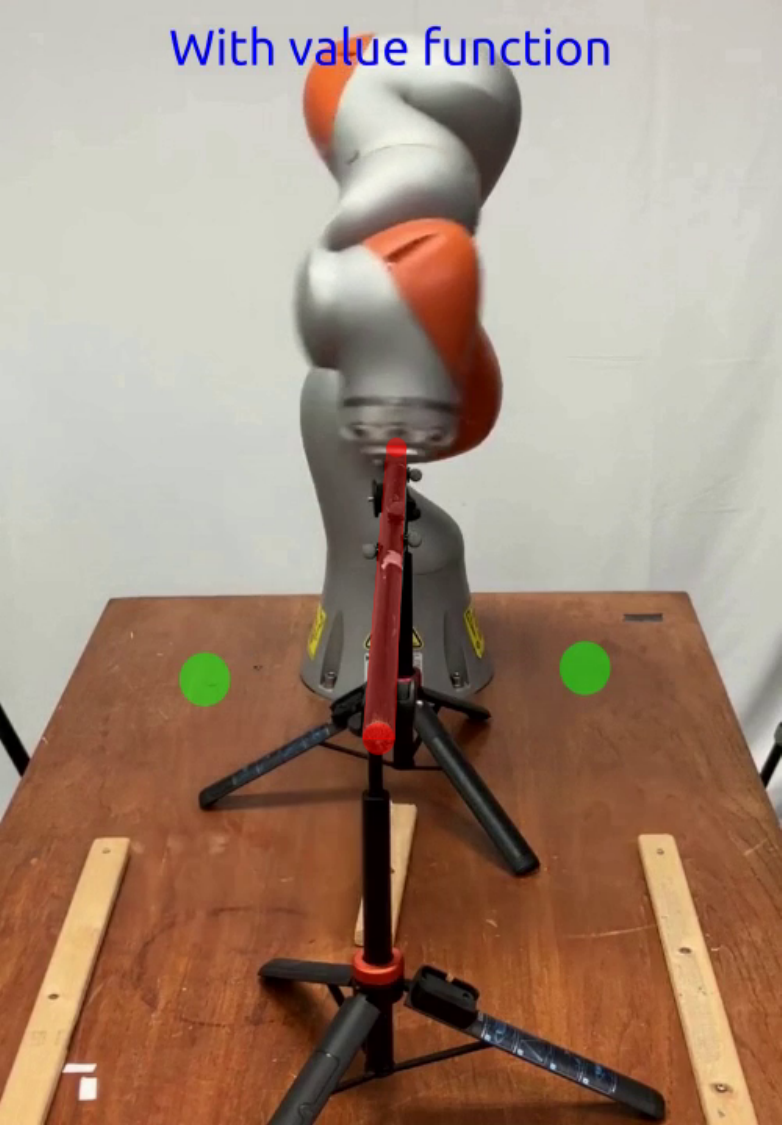}&
  \includegraphics[width=.12\textwidth, height=.12\textheight]{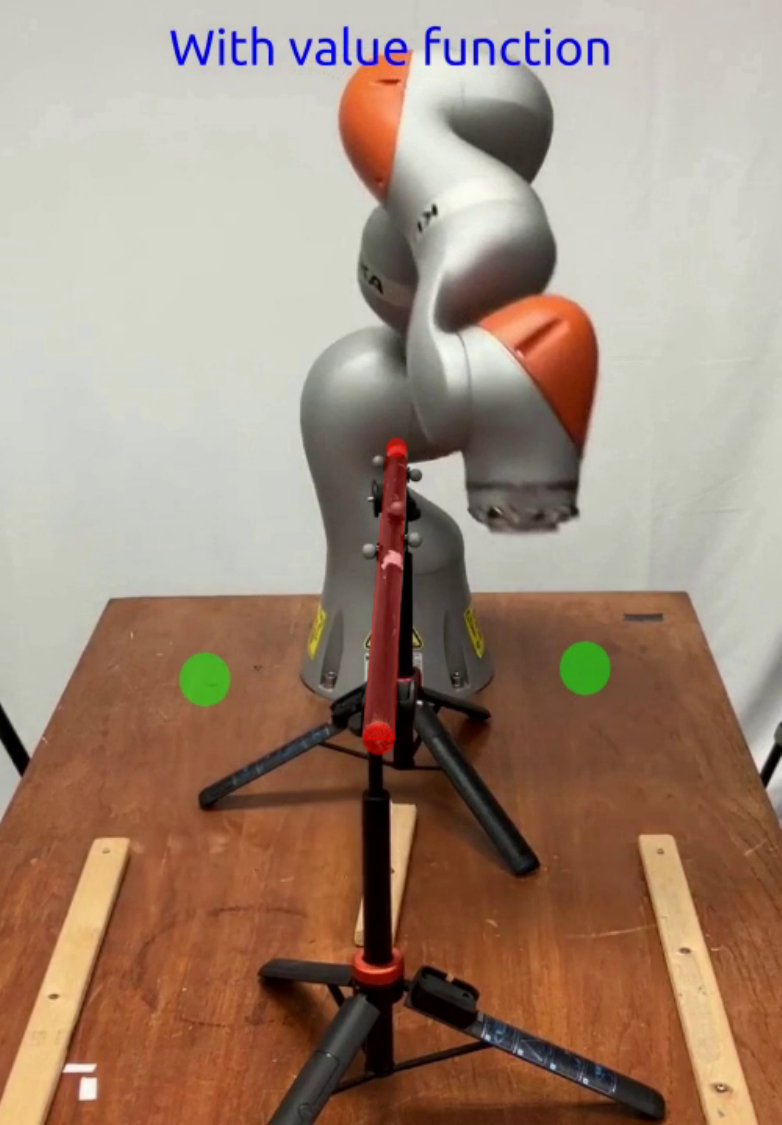}&
  \includegraphics[width=.12\textwidth, height=.12\textheight]{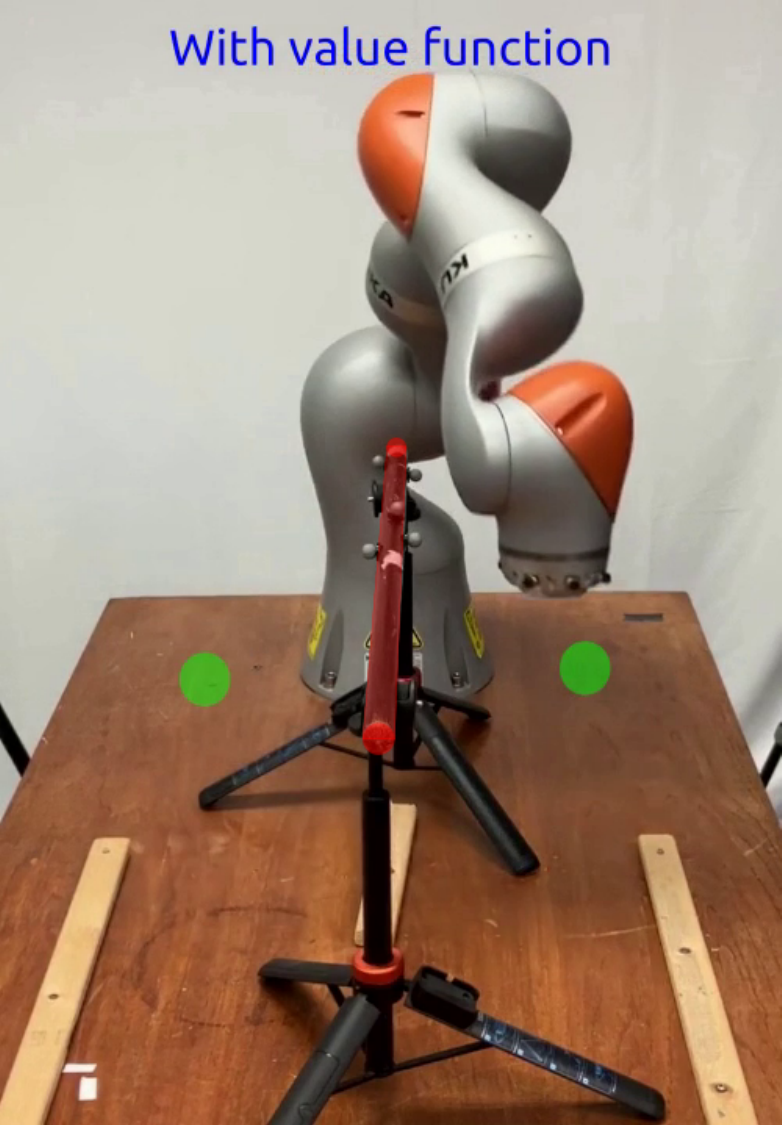}&
  \includegraphics[width=.12\textwidth, height=.12\textheight]{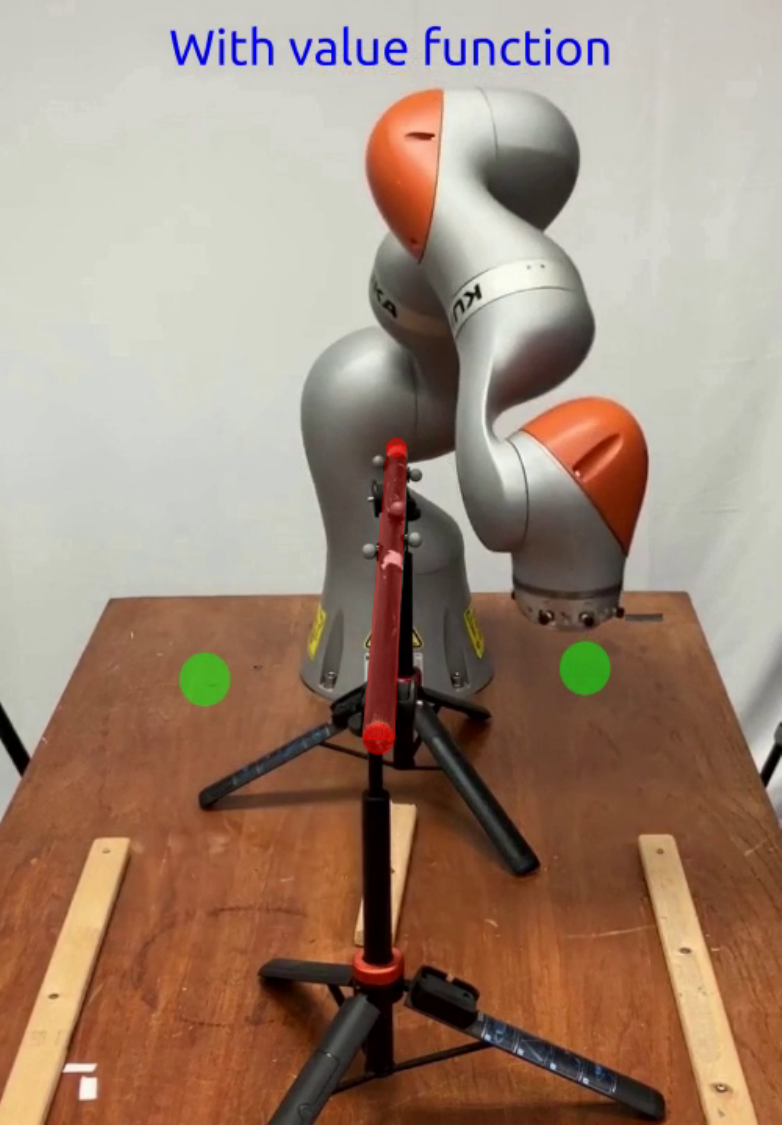}&
  \includegraphics[width=.12\textwidth, height=.12\textheight]{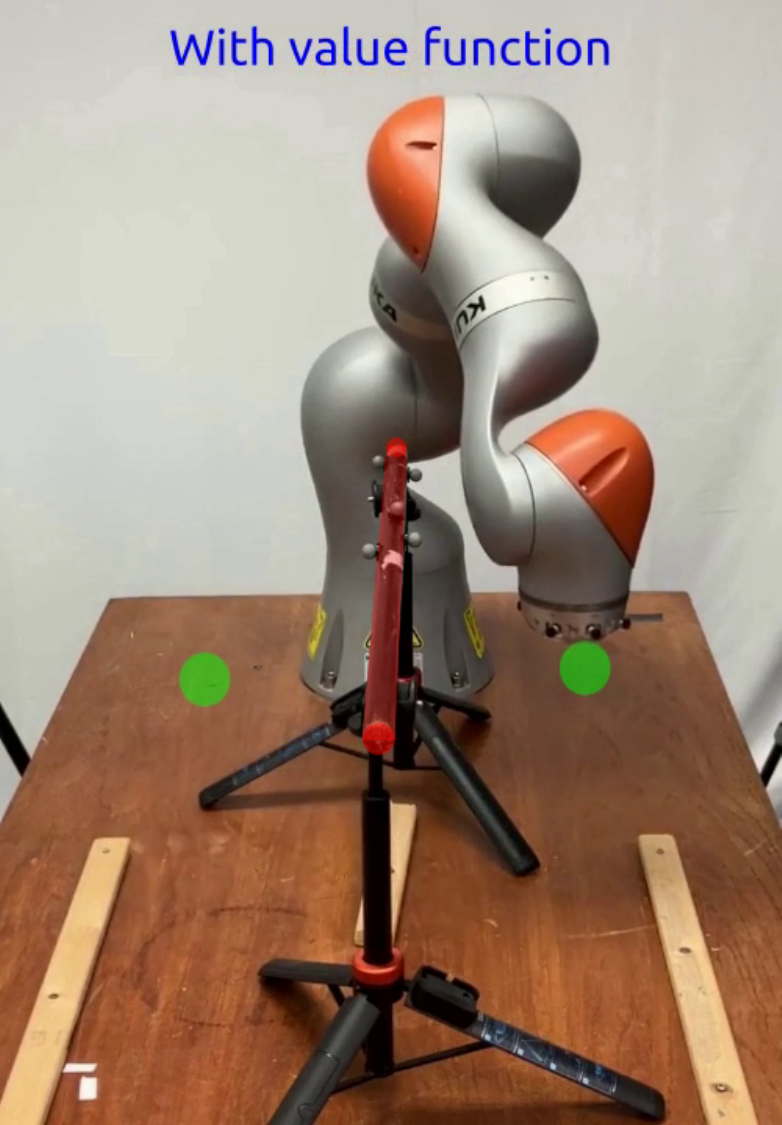}&
  \includegraphics[width=.12\textwidth, height=.12\textheight]{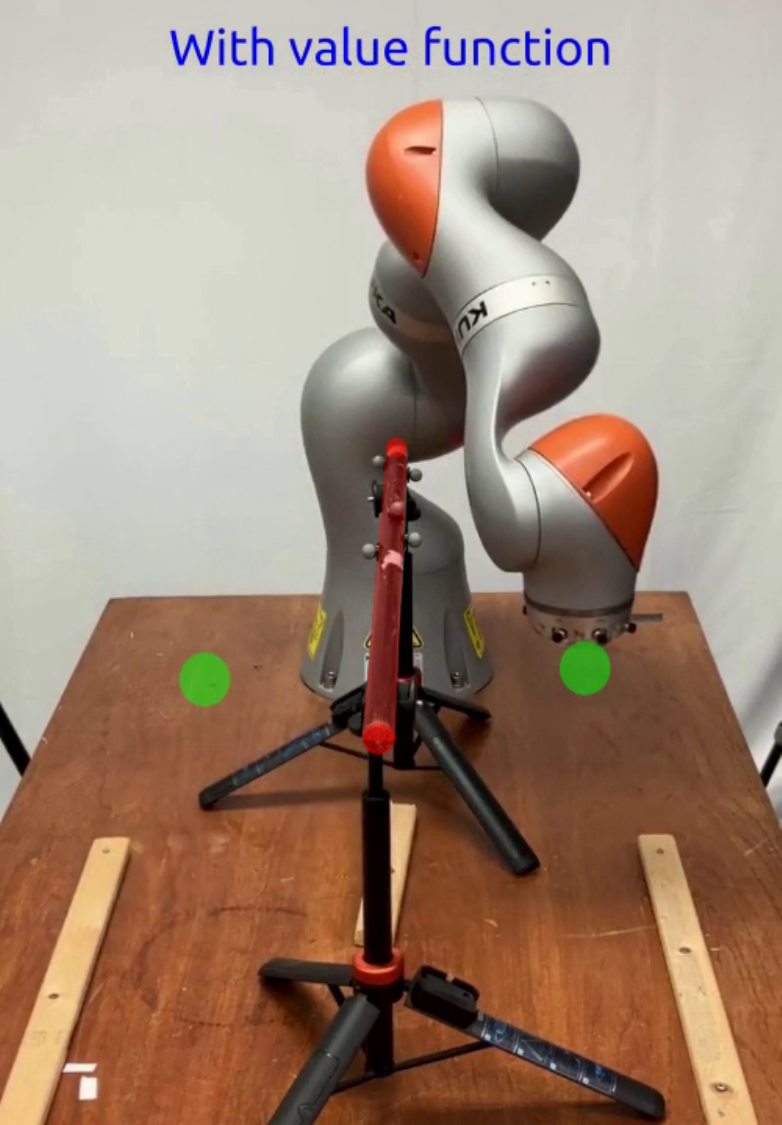}\\
  \includegraphics[width=.12\textwidth, height=.12\textheight]{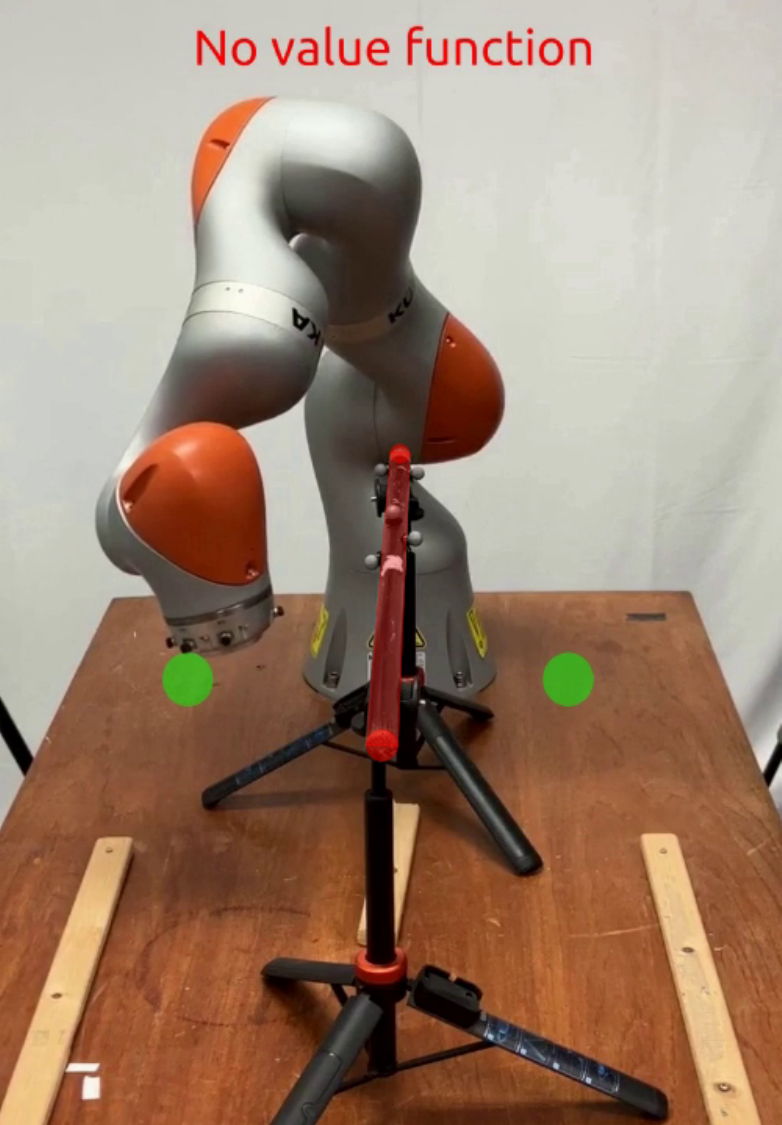}&
  \includegraphics[width=.12\textwidth, height=.12\textheight]{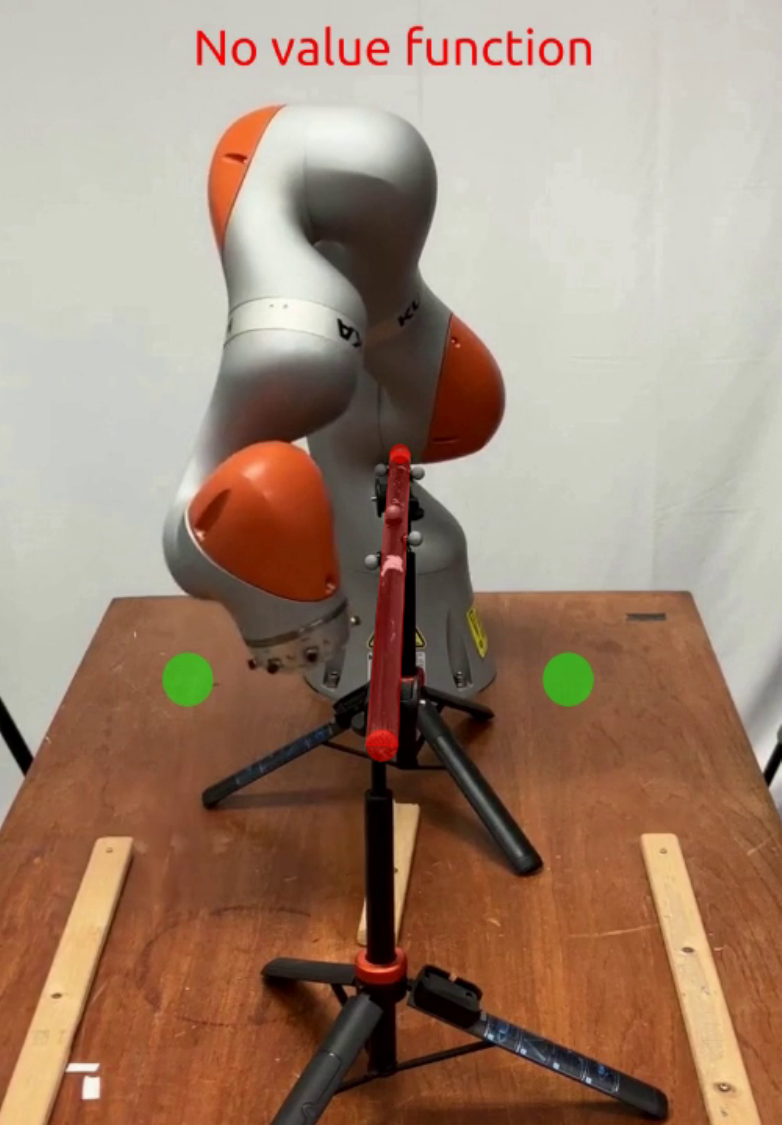}&
  \includegraphics[width=.12\textwidth, height=.12\textheight]{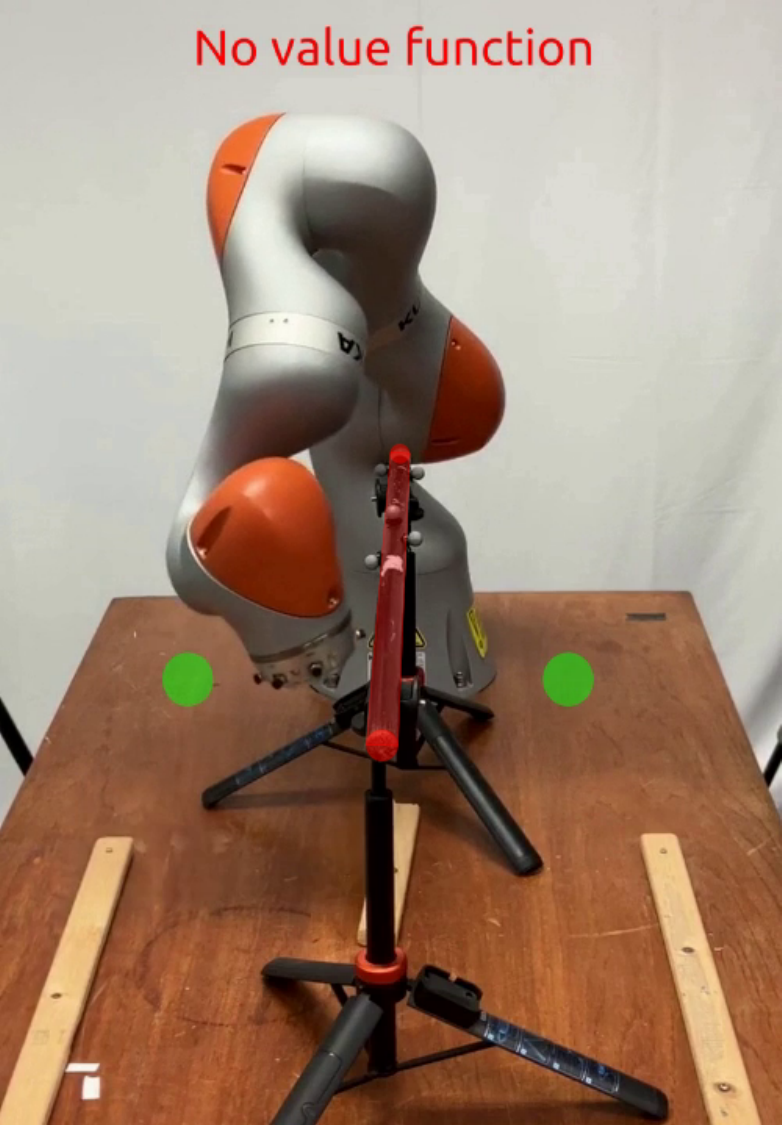}&
  \includegraphics[width=.12\textwidth, height=.12\textheight]{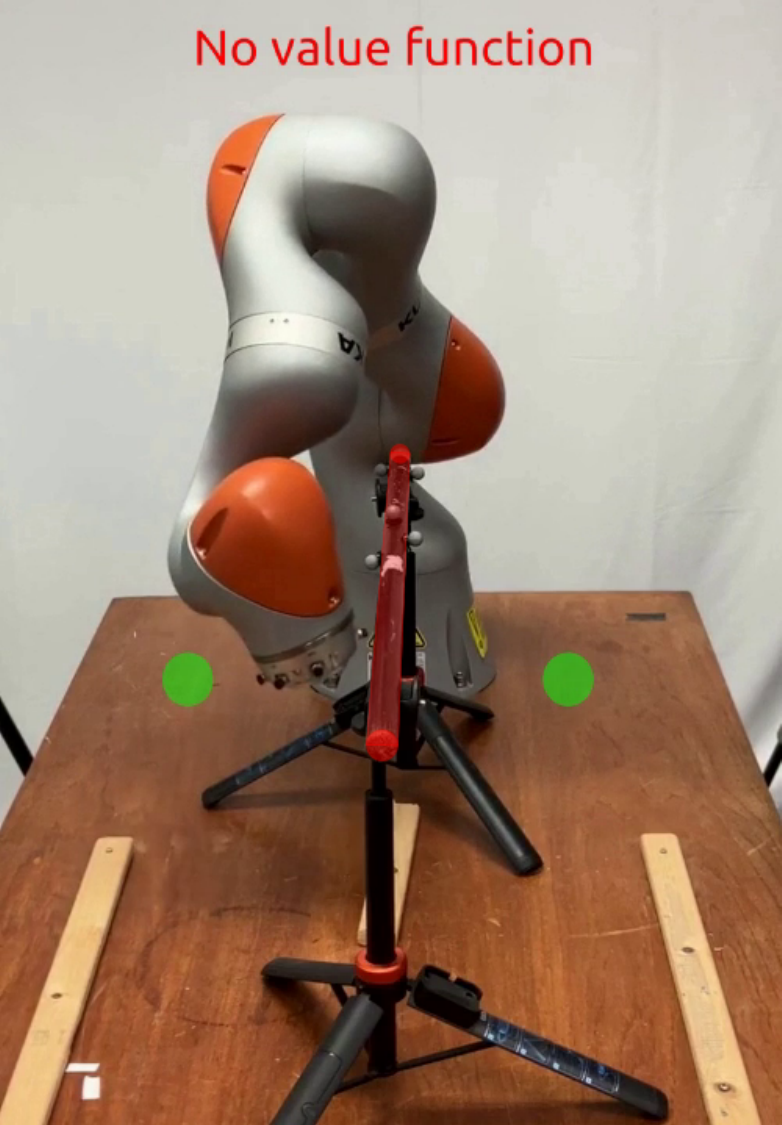}&
  \includegraphics[width=.12\textwidth, height=.12\textheight]{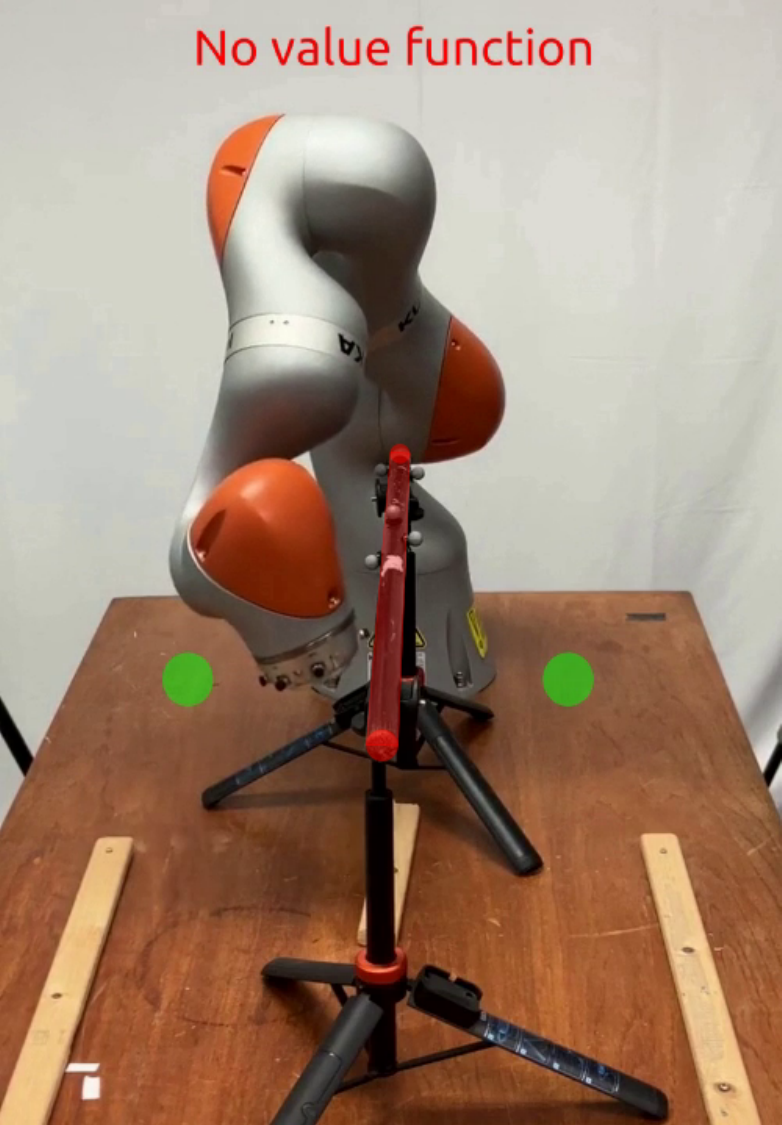}&
  \includegraphics[width=.12\textwidth, height=.12\textheight]{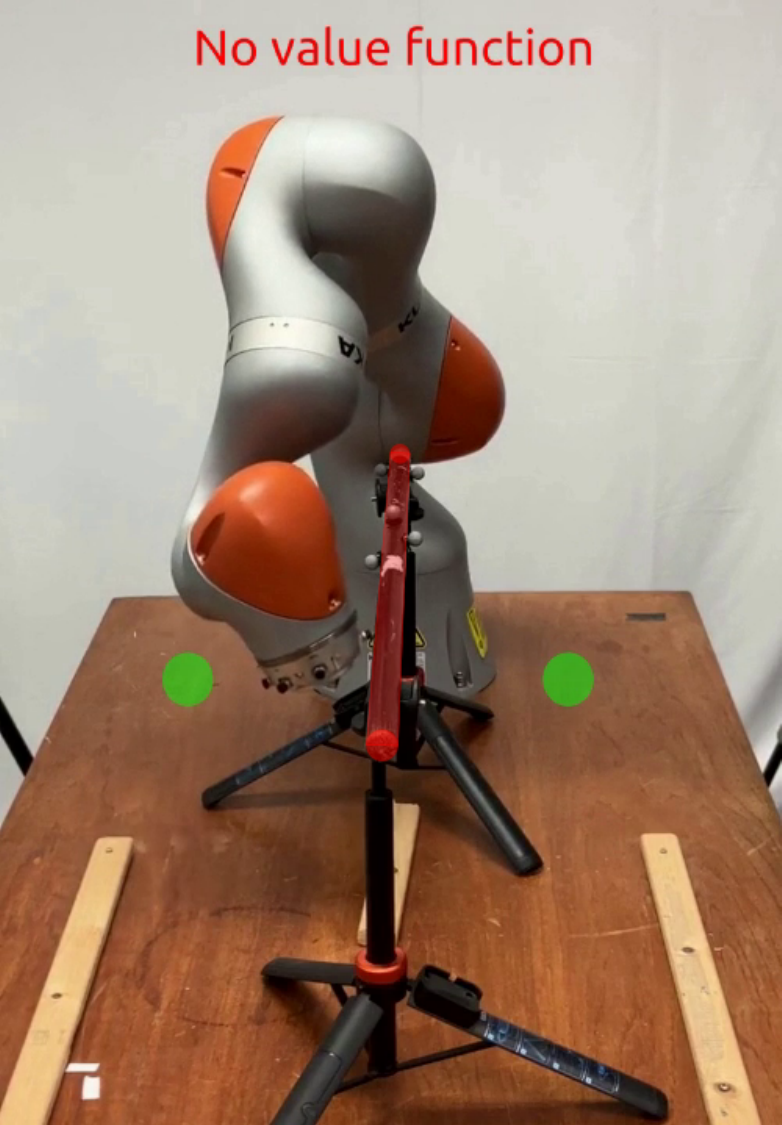}&
  \includegraphics[width=.12\textwidth, height=.12\textheight]{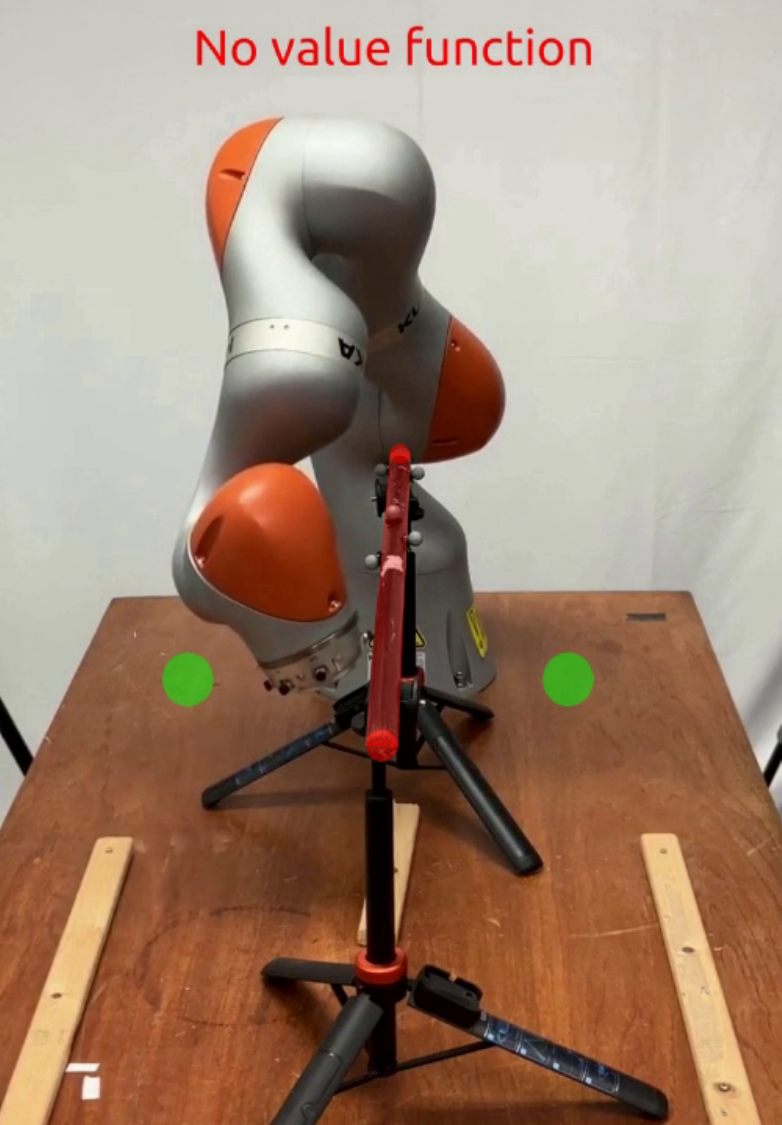}&
  \includegraphics[width=.12\textwidth, height=.12\textheight]{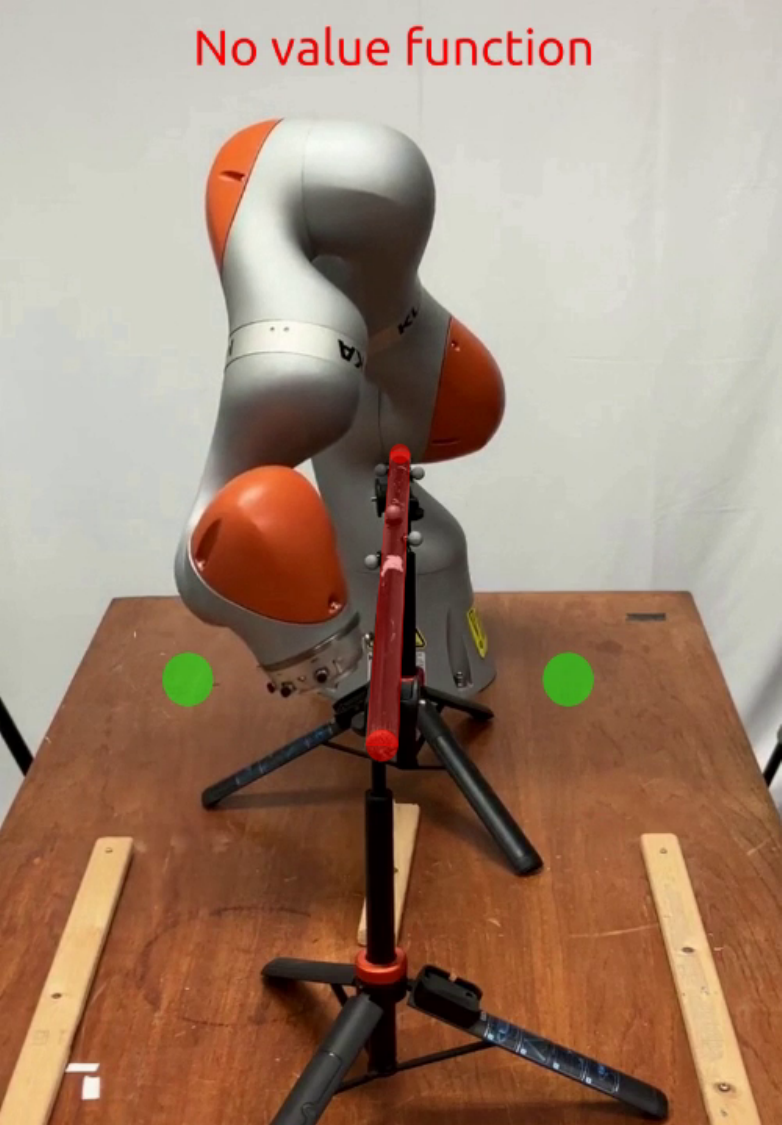}
  
\end{tabular}
    \caption{Snapshots of pick-and-place task with static obstacle avoidance for the default MPC without value function (bottom) and the proposed MPC with value function (top). The green dots represent the end-effector targets that must be reached alternatively while avoiding collision with the black rod placed in the center (highlighted in red).}
    \label{fig:snap}
\end{figure*}
We compared the performance of the proposed approach against the default MPC on a pick-and-place task with a static obstacle. The robot must alternatively reach two end-effector positions while avoiding collision with a fixed rod laying in between the targets. Figure~\ref{fig:snap} shows that the MPC with value function reaches the target (by finding a path moving the end-effector \textit{above} the rod) while the default MPC remains stuck in a local minimum (trying to go \textit{underneath} the rod). This behavior can be further understood by looking at the total cost after $10$ seconds. With the value function, the final cost is $213$ while it reaches $1311$ without it.
The MPC with value function achieves a lower cost since it eventually reaches the target. Interestingly, this controller initially increases its cost faster than the default MPC. This is due to higher velocity and torque regularization cost residuals during the obstacle avoidance motion. Hence, this experiment shows the ability of the proposed controller to avoid local minima thanks to the value function. Indeed, the default MPC remains stuck in a local minimum and must trade off the task completion against constraint satisfaction. In contrast, our approach is able to both achieve the task \textit{and} avoid collision with the rod by choosing a different path that initially increases the cost. It is important to remind that both controllers use the same warm-start, the only difference is the terminal cost used. 

\subsubsection{Target tracking with static obstacle}
\begin{figure}
    \centering
    \includegraphics[width=\linewidth]{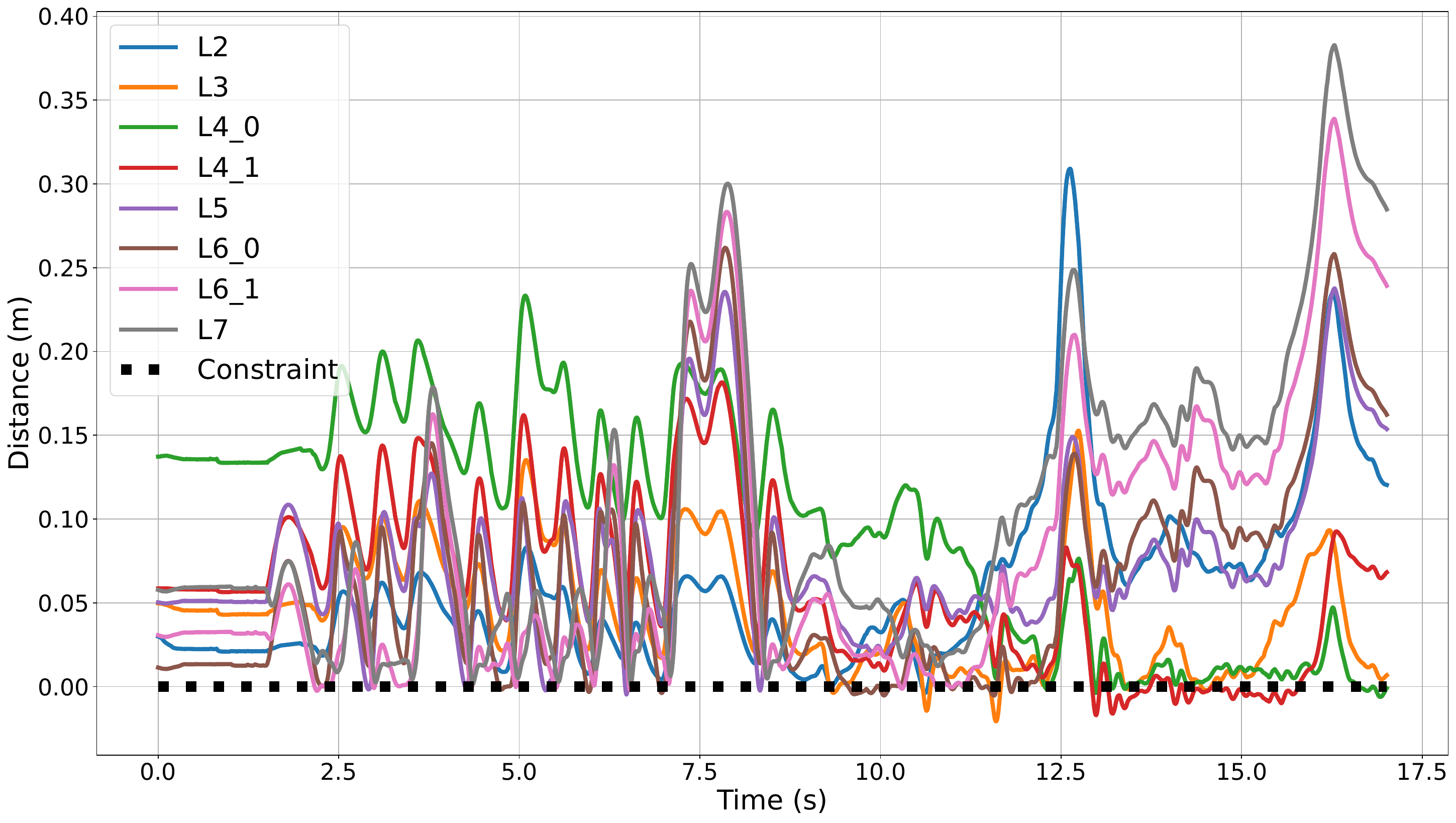}
    \caption{Target tracking with static obstacle. Collision signed distance between the robot's capsules and the obstacle.}
    \label{fig:exp2_distance}
\vspace{-0.4cm}
\end{figure}

In this experiment, we show the ability of our approach to track a moving target while satisfying obstacle avoidance constraints. We use a small cube tracked by the motion capture system to define a moving target. The accompanying video~\footnote{\url{https://youtu.be/CruTx2CvcFQ}} shows how the controller can go around the obstacles when the cube is moved from one side to the other of the obstacle. Figure~\ref{fig:exp2_distance} depicts the constraint satisfaction between the rod and the capsules. 
The negative values indicate small constraint violations which can be explained by the fact that the solver does not always reach full convergence. In practice, the capsules are slightly larger than the robot, and the constraint violations are small enough for the robot to avoid collisions on the hardware.

\subsubsection{Target tracking with dynamic obstacle} Lastly, we illustrate in the video the ability of the method to deal with out-of-distribution orientation of the rod as well as unexpected disturbances.
Intuitively, far from the training distribution, the approximated value function is not very meaningful and cannot provide a way to avoid the obstacle. In fact, it can be seen that the controller is no longer able to systematically avoid local minima due to the obstacle. However, the hard constraints are satisfied. 

\section{DISCUSSION}

While value iteration is guaranteed to converge~\cite{bertsekas2015value}, this relies on the assumption that the minimization in Problem~\eqref{main_eq} is global. However, we use a gradient-based solver which can be subject to local minima. Intuitively, the larger $T$ is, the more the solver is prone to local minima. In practice, we found that tuning appropriately $T$ was enough to get VI to converge. In the end, the choice of $T$ regulates the trade-off between the efficiency of the local solver and the global property of value iteration. Lastly, we have investigated the use of various random warm-starts to search for the global solution; however, the convergence speed-up did not compensate for the additional computational time. Nevertheless, this remains an interesting direction to explore further.

Another limitation of our method is that gradient-based solvers require a smooth neural network. This constrains the network architecture and training parameters. More specifically, the SQP solver could not handle a network with ReLU activation and required an appropriate tuning of the weight decay. Without weight decay, the network would overfit and the solver's number of iterations would diverge. To circumvent the issue, it would be interesting to investigate the use of zero-order methods, which have recently shown promising results~\cite{li2024drop, xue2024full}.

One of the key assumptions of the work is the tractability of the feasible set $\Omega$. Although this assumption encompasses a wide set of problems, it would be interesting to study how to generalize to any type of constraint. A naive approach could be to first approximate $\Omega$ with other methods such as \cite{djeridane2006neural, bansal2021deepreach} and then apply value iteration. However, it would be interesting to combine those two steps.

This work focuses on the non-discounted setting because this is the original MPC formulation that can guarantee stability~\cite{mayne2014model, grune2017nonlinear}. Arguably, using a non-discounted setting requires a cautious design of the problem as it is crucial to ensure that the goal states achieve zero cost. Furthermore, while \cite{heydari2014revisiting, bertsekas2015value} proved the convergence of value iteration in the non-discounted setting, it is still not clear how to guarantee convergence while approximating the value with neural networks. In contrast, popular RL algorithms usually use a discount factor~\cite{sutton2018reinforcement}, It would be interesting to study the impact of the discount parameter during training. However, it is not clear if the stability guarantees of the infinite horizon~\cite{grune2017nonlinear} will be preserved in that setting. 

In the manipulator experiment, one limitation is that our training assumes that the goal and the obstacle do not move. If we knew their dynamics, we could add them to the state space and apply our method. However, this could make the set $\Omega$ intractable. 
Lastly, for more complex problems, there might be a trade-off between the neural network expressivity and inference time: a large network might increase the accuracy but also slow down the evaluation time, which may in turn limit the horizon length that can be used during deployment. However, in our experiments, the network architecture yielding the best approximation was not very large and its inference time was comparable to that of the model. As a result, we did not face this trade-off in practice.

\section{CONCLUSIONS}

We have introduced a way to combine constrained TO with RL by using a learned value function as a terminal cost of the MPC. We have demonstrated the benefits of the approach on a reaching task with obstacles on an industrial manipulator. In contrast to traditional MPC, by approximating an infinite horizon OCP, the method can avoid complex local minima. Furthermore, in contrast to RL, the online use of TO allows us to gain accuracy as it can leverage online the model of the robot. Furthermore, the proposed method allows us to handle hard constraints and out-of-distribution states. Future work will investigate how to extend this work to problems involving contacts such as locomotion and manipulation.

\addtolength{\textheight}{-0.4cm}   




\bibliographystyle{IEEEtran}
\footnotesize

\bibliography{references}

\end{document}